\documentclass[11pt]{article}
\usepackage{acl2016}
\usepackage{times}
\usepackage{url}
\usepackage{latexsym}
\usepackage{color}

\usepackage{amsmath}
\usepackage{amssymb}
\usepackage{amsfonts}
\usepackage{tabu}
\usepackage{xcolor}
\usepackage{graphicx}
\usepackage{bm}
\usepackage{multirow}

\DeclareMathOperator*{\argmin}{arg\,min}

\newcommand\R{\mathbb{R}}

\newcommand{\mat}[1]{\boldsymbol{\mathbf{#1}}}

\newcommand{\bb}[1]{\mathbb{#1}}

\aclfinalcopy % Uncomment this line for the final submission
\title{
	Bilingual Dictionary Induction for Bantu Languages}

\author{Ndapa Nakashole \\
 Computer Science and Engineering  \\
  University of California, San Diego\\
  La Jolla, CA 92093 \\
  {\tt nnakashole@eng.ucsd.edu} \\}

\date{}

\begin{document}
\maketitle
\begin{abstract}
We present a method for learning  bilingual
translation dictionaries between English and Bantu languages.
%Our method only requires training data in the form of partial dictionary  for one the Bantu languages.
We show that exploiting the grammatical structure common to  Bantu languages enables    bilingual dictionary induction for  languages where training data is unavailable.
%We show that
%high-precision lexicons can be learned in a variety
%of language pairs and from a range of
%corpus types.

%%%% FOR CAMERA READY, ADD PLOT OF WAMBO AND NDONGA EMBEDDINGS

%%%%NB

\end{abstract}

\section{Introduction}
\begin{figure}[t]
  	\centering
  	\vspace{-0.1cm}
  	\includegraphics[width=0.6\columnwidth]{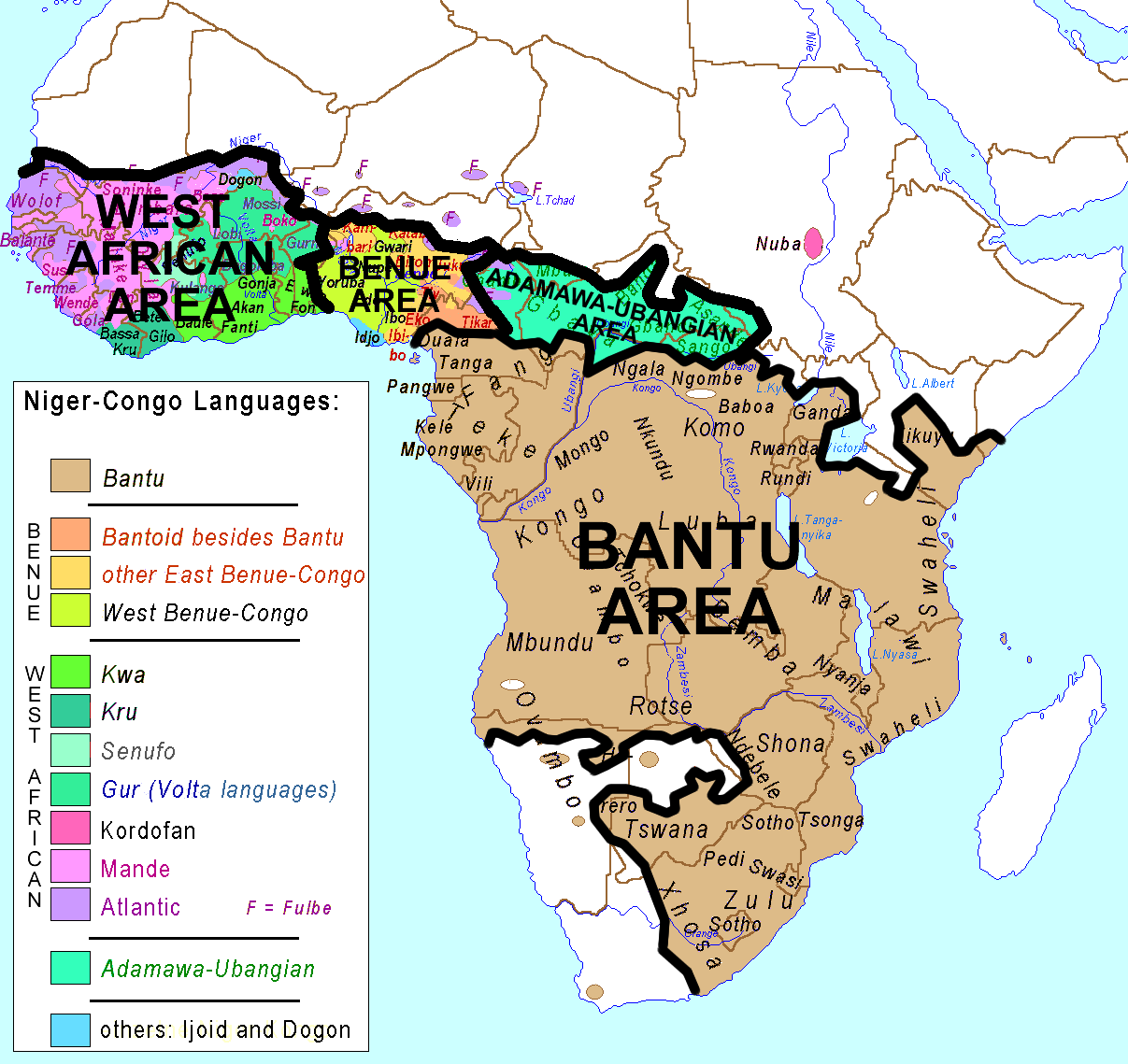}
  	\vspace{-0.1cm}
  	\caption{ Bantu languages are spoken over much of central and southern  Africa.}
  	% of section \ref{sec:patterngenerated}. For readability, each data point is shown as the last three characters of  the concept name.}
  	\label{fig:bantumap}
  \end{figure}

 Bilingual dictionaries mostly  exist for  resource-rich language pairs, for example,  English-German and   English-Chinese \cite{Koehn2002,conf/acl/HaghighiLBK08,DBLP:journals/corr/AmmarMTLDS16,DBLP:conf/eacl/FaruquiD14}. Such dictionaries are useful for many natural language processing (NLP) tasks including statistical machine translation,  cross-lingual information retrieval, and  cross-lingual transfer of  NLP models such as those for  part-of-speech tagging and dependency parsing  \cite{DBLP:conf/naacl/TackstromMU12,DBLP:conf/acl/GuoCYWL15,DBLP:conf/naacl/GouwsS15}.
In this paper, we consider the task of bilingual dictionary induction for English and Bantu languages.  Bantu languages are a family of over 300\footnote{Between 300 and 600, depending on where the line is drawn between language and dialect.}  mutually intelligible languages  spoken over much of central and southern Africa, see the map\footnote{ Image from http://getyours-it.nl/a-culturu/afrikaanse-stammen/bantu-stammen} in  Figure \ref{fig:bantumap}  \cite{guthrie1948classification,nurse2003towards}.   About a third of  Africans speak  a Bantu language as their native language~\cite{nurse2001survey}. The most widely spoken, with over 10 million speakers\footnote{https://www.ethnologue.com} each, include 
Swahili (Kenya, Uganda), Shona (Zimbabwe), Zulu (South Africa), and Kinyarwanda (Rwanda, Uganda).

 As with other low resource languages, labeled data for Bantu languages is scare.
We seek to exploit the Bantu grammar structure  to mitigate  lack of  labeled data. 
%Given that some Bantu languages have more resources than others.
 More specifically, we ask the following question: given   a small bilingual dictionary  between English and one Bantu language, $ L_{bantu1}$,   can we 1)  infer missing entries in the $English - L_{bantu1}$  dictionary 2) generate a new bilingual  dictionary $English- L_{bantu2}$ for another Bantu language for which labeled data is unavailable.
%  For case 2, the only form of supervision comes indirectly from the partial dictionary, $English - L_{bantu1}$. 
 To answer this question we propose an approach based on distributed representations of words \cite{DBLP:journals/jair/TurneyP10,MikolovSkip}. The first step is to create a vector space for each language,  derived from a text corpus for the language. Notice that these text corpora need not be aligned. The second step is to perform dictionary induction by learning 
a linear projection, in the form of a matrix, between language vector spaces \cite{DBLP:journals/corr/MikolovLS13,DBLP:journals/corr/DinuB14,DBLP:conf/acl/LazaridouDB15}. Our key insight for Bantu languages is that one can create a single vector space for them, obviating the need for learning a  projection matrix for each Bantu language.  This means we only need to learn  a single projection matrix,  for inducing multiple English to Bantu bilingual dictionaries,  using 
the small bilingual dictionary  $English - L_{bantu1}$.  Additionally,  we modify the corpus corresponding to $L_{bantu2}$ to have a greater vocabulary intersection with $L_{bantu1}$. This step is inspired by the  extensive use of bases and affixes, common to Bantu languages.  Words with the same meaning often differ only in the affixes  with the base being   similar or the same. We therefore use edit distance to replace some fraction of the words of $L_{bantu2}$  with similar  words in $L_{bantu1}$.

 \paragraph{Contribution.} 
 1). \textit{Unsupervised Bantu language dictionary induction}: To the best of our knowledge,  our work is the first effort
 to create bilingual dictionaries for Bantu languages using unsupervised machine learning methods.
 2) \textit{Data:} We collect corpora for seven  Bantu languages. Having had access to a first language speaker of two Bantu languages, we obtained labeled  which we make available along with the corpora, for further research into NLP for Bantu languages.
 3) \textit{Dictionary induction almost  from scratch:} We propose a method for dictionary induction that only requires training data in one of the Bantu languages.  Our experiments  show the  potential of our approach.
 
%We propose an approach that does this. We test our approach on two Bantu languages.

\section{Approach}
% Talk about embeddings. % refs from 
% Bilingual Word Embeddings from Non-Parallel Document-Aligned Data
%Applied to Bilingual Lexicon Induction
\subsection{Distributed Representation}
Distributed representations of words, in the form of real-valued vectors,
encode word semantics based on  collocation of  words in text \cite{DBLP:journals/jair/TurneyP10,MikolovSkip,DBLP:journals/corr/AmmarMTLDS16}.
Such vector representations have been shown to  improve performance of various NLP tasks
including semantic role labeling, part-of-speech tagging, and named entity recognition \cite{DBLP:journals/jmlr/CollobertWBKKK11}.
%2008; ). Many models have been introduced for learning such word embeddings (Mikolov et al., 2013c, Turney
%and Pantel, 2010, Pennington
%et al., 2014). 
In this work we use the skip-gram model with negative sampling to generate word vectors
\cite{MikolovSkip}. It  is one of the  most competitive methods
for generating  word vector representations,  as  demonstrated by results on a various semantic tasks \cite{conf/acl/BaroniDK14,DBLP:journals/corr/MikolovLS13}.

\subsection{Bilingual Dictionary Induction}
To induce a bilingual dictionary for a pair of languages, we use the projection matrix approach  \cite{DBLP:journals/corr/MikolovLS13,DBLP:journals/corr/DinuB14,DBLP:conf/acl/LazaridouDB15}. It takes as input a small bilingual dictionary containing pairs of translations  from the source language  to the target language.   Training data is comprised  of  vector representations  of word pairs $D^{tr}=\{x_{i}, y_{i}\}^{m}_{i=1}$, where $x_{i} \in
\mathbb{R}^{s}$ is the vector for word $i$ in the
source language, and $y_{i} \in \mathbb{R}^{t}$ is the vector for its translation in the target language. At test time, we predict the target word translations for new source language  words,  $D^{te}=\{x_{j}\}^{n}_{i=1}$  where $x_j  \in  \mathbb{R}^{s}$.
In our case, the source language is  a Bantu language and  the target language is English.

This  approach assumes that there is linear relationship between the two vector spaces.
Thus, the learning problem is to find a matrix $\mat{W}$ that maps a source language word vector $x_i$ to the vector of its translation $y_i$ in the target language.  As in  \cite{DBLP:journals/corr/DinuB14}, we  use an l2-regularized least squares error to learn the projection matrix  $\mat{W}$.
\begin{equation}
\hat{\mat{W}} = \argmin_{\mat{W} \in \bb{R}^{s \times t}}||\mat{X}\mat{W}-\mat{Y}||_F + \lambda||\mat{W}||
\label{eq:obj1}
\end{equation}
where \mat{X} and \mat{Y} are matrices  representing the  source  and  target vectors in the training data, respectively.
For a new  Bantu  word whose   vector
representation  is $x_j \in  \mathbb{R}^{s}$, we  map it to English  by computing
$\hat{y_j} = \hat{\mat{W}}x_j$ where $\hat{y_j} \in \mathbb{R}^{t}$ , and then finding  the English word  whose vector representation is closest to $\hat{y_j}$, as measured by  the  cosine similarity distance metric.

\subsection{Bantu Language Structure}
The word ``Bantu" is derived from the word for ``people", which has striking similarities in many  Bantu languages \cite{guthrie1948classification,nurse2003towards}. In   Zulu   (South Africa) ``\textit{abantu}" \footnote{Ubuntu is Zulu for  humanity or  the essence of being human.} means  \textit{people};  in Swahili  (Kenya, Uganda) ``watu" ;  in Ndonga  (Namibia) ``aantu";  in Sesotho (Lesotho) ``batho"; in Herero (Namibia) ``ovandu"; and  in Kwanyama (Namibia, Angola)  ``ovanhu".  
 it is often used in the philosophical sense " \footnote{South African Nobel Laureate Archbishop Desmond Tutu describes Ubuntu as:
	"It is the essence of being human. It speaks of the fact that my humanity is caught up and is inextricably bound up in yours. ``A person with Ubuntu is open and available to others, affirming of others, does not feel threatened that others are able and good, based from a proper self-assurance that comes from knowing that he or she belongs in a greater whole and is diminished when others are humiliated or diminished, when others are tortured or oppressed. "}.
While Bantu languages may differ  in vocabulary, in some cases quite substantially,  they share the same grammatical structure. A prominent aspect of the grammar of Bantu languages is the extensive use of  bases and affixes. For example, in the country of Botswana, from the base Tswana,  the people of Botwana are the Batswana, one person is a Motswana, and the language is Setswana.  We seek to exploit this property by performing edit distance corpus modifications before learning the projection matrix. %approach to dictionary induction.

 	\begin {table}[t]
 	\centering
 	\resizebox{\columnwidth}{!}{%
 		\begin{tabular}{|l|l|l|l|}
 			\hline
 			&  & Voca-& KW \\
 			& Tokens &  bulary &  vocab $\cap$\\
 			\hline
 			KW-kwanyama  & 732,939& 33,522 & 33,522\\
 			ND-Ndonga  & 732,939& 33,522 & 3,769\\
 			SW-Swahili & 694,511 &  49,356& 173  \\
 			KK-Kikuyu & 718,320 & 53623 &  126 \\
 			SH-Shona & 570,778 & 64,073 &222 \\
 			CW-Chewa & 669, 352 & 53148 & 206 \\
 			TS-Tswana & 101,1750 & 23,384 &126\\
 			\hline
 			EN-English & 875,710 & 13,897 & 26 \\
 			IT-Italian & 829,709 & 29,534 & 78\\
 			\hline
 		\end{tabular}
 	}
 	\label{tbl:corpora}
 	\caption{Corpora crawled from Bible.com:  seven Bantu   and two  Indo-European languages.}
 	\end {table}
 	
\subsection{Single Projection Matrix}
We hypothesize that   we only need to learn one projection matrix,  $\mat{W}$ in Equation~\ref{eq:obj1}.
Our labeled data is a small bilingual dictionary   $English - L_{bantu1}$, between English and a Bantu language $L_{bantu1}$.  We would like to be able to  infer missing entries in  the $English - L_{bantu1}$ dictionary, and to  generate a new  dictionary, $English - L_{bantu2}$,  a language pair for which  labeled data is unavailable.  The core idea is to  create only one vector space for the two Bantu languages. First we generate  a lexicon, $lex_{b2}$,  containing  words that appear in  the corpus of $L_{bantu2}$. Next,  for each $w \in lex_{b2}$ we find all  words in $lex_{b1}$,  the lexicon of language $L_{bantu1}$,  whose edit  distance  to $w$  is a small value $ \Phi$\footnote{In the experiments we use $\Phi = 1$.}. Thus, each word $w \in lex_{b2}$ has a  list of  words from $lex_{b1}$,  $S_w = \{ w1 \in Lex_{b1}: editditance(w, w1) <= \Phi \}$. We then go through corpus  $C_{bantu2}$ and with  probability $\Pi$\footnote{In the experiments we use $\Pi = 0.5$.},  we replace word $w \in lex_{b2}$ with  one of its cross-lingually similar words in $S_w$, random selection is used to pick the replacement word from $S_w$. This  process creates an updated corpus $\hat{C}_{bantu2}$.   Concatenating the Bantu corpora gives:
%\begin{equation}
$C_{bantu}= C_{bantu1} \cup \hat{C}_{bantu2}$\label{eq:bantucorpus}.
%\end{equation}
Applying the skip-gram model to  $C_{bantu}$, generates word vectors, every  $w_i \in C_{bantu}$ has a  vector  $x_i \in \R^{s}$. 
%We also generate target vector space embeddings  for the English words using an English corpus such that every English word has  a corresponding vector representation $y_i \in \R^{t}$.
For English,   we use  the 300-dimensional pre-trained vectors trained the Google News dataset \footnote{https://code.google.com/archive/p/word2vec}, so that every English word has  a  vector  $y_i \in \R^{t}$. 
Finally, we  $\mat{W}$ using the 
 training data, $D^{tr}=\{x_{i}, y_{i}\}^{m}_{i=1}$. At test time, we predict the target word translations for unseen Bantu  words, $D^{te}=\{x_{j}\}^{n}_{i=1}$, which can either be in language  $L_{bantu1}$ or $L_{bantu2}$.

 \begin {table}[t]
 \centering
 \resizebox{\columnwidth}{!}{%
 	\begin{tabular}{|l|l|l|}
 		\hline
 		\textbf{English -EN} &\textbf{Kwanyama-KW} & \textbf{Ndonga-ND} \\
 		\hline
 		people	& ovanhu &	aantu \\
 		return &  aluka	 & galuka \\
 		child	& okaana& 	omunona	\\
 		things &	oinima &	iinima \\
 		ears& 	omakutwi	 & omakutsi \\
 		women	& ovakainhu	& aakiintu \\
 		tongue &	elaka &	elaka \\
 		darkness&	omulaulu &	omilema \\
 		feet &	eemhadi	& oompadhi \\
 		sins	& omatimba	& oondjo \\
 		
 		\hline
 	\end{tabular}
 }
 \label{tbl:examples}
 \caption{Examples of English, Kwanyama, and Ndonga translations}
 \end {table}

 	\begin {table}[t]
 	\centering
 	\resizebox{\columnwidth}{!}{%
 	\begin{tabular}{|l|l|l|l|}
 		\hline
 		& Target &Training & \\
 		& Vocabulary &  Dictionary &  Test\\
 		\hline
 		EN-KW & 33,522 & 2,142& 107\\
 		EN-IT &  200,000& 5,000 &1,500  \\
 		EN-ND & 32,026 & 0 &  104 \\
 		\hline
 	\end{tabular}
 }
 	\label{tbl:labeddata}
 	\caption{Training and test data for dictionary induction.}
    
 	\end {table}

\section{Experimental Evaluation}
\paragraph{Data.} We crawled  Bible.com for  bibles of 9 languages,  7 Bantu  and 2 Indo-European,  Italian and English. The latter were used for comparison. Corpora statistics are shown in Table 1.
% \ref{tbl:corpora}.
  In our experiments, we focused  on Kwanyama,  spoken Namibia and Angola, as we had access to a first language speaker who could annotate data. The last column of Table  1
%\ref{tbl:corpora}
 shows the vocabulary intersection between Kwanyama and other  languages. The language with the most words in common with  Kwanyama is Ndonga,  spoken in Namibia, with an 11\% vocabulary overlap. As expected, the Indo-European languages  overlap the least with Kwanyama and the few words they have in common have different meanings. For example: The word ``male" in English refers to gender, in  Kwanyama  ``male" means ``tall" or ``deep".  Our Kwayama first language speaker  also has five years of formal training  in Ndonga, which is  a dialect of the same language as Kwanyama. 
% Given the 11\% vocabulary overlap, Kwanyama and Ndonga are a good candidate for testing our proposed single projection matrix approach. 
 We therefore focus on these two Bantu languages in our experiments. Table 2 shows some examples of English, Kwanyama and Ndonga translations.  Details of the training  and test  data are shown in Table 3. For all languages,  we used 300-dimensional word vectors.
 % \ref{tbl:labeddata}.

	\begin {table}[t]
	\centering
	\resizebox{\columnwidth}{!}{%
	\begin{tabular}{|l|l|l|l|l|}
		\hline
		& \textbf{P@1} &\textbf{P@5}  &\textbf{P@10} & \textbf{RD}\\
		& & & & =0.10\\
		\hline
		EN-KW & \textbf{0.30 }& \textbf{0.56} & \textbf{0.58} &  \textbf{0.86 }\\
		EN-KW (RD)  &  0.00 & 0.00 &  0.00 & 0.10 \\
		\hline
		EN-IT  & \textbf{0.34} & \textbf{0.48} & \textbf{0.54} & \textbf{0.94}\\
		EN-IT (RD) & 0.00 & 0.00 & 0.00 &0.10\\
		\hline
		EN-ND (J-IT) & 0.00 & 0.00 & 0.00 & 0.39 \\
		EN-ND (J-KW) & 0.07& 0.16 & 0.18 & \textbf{0.63} \\
		EN-ND (J-KW-R)  & \textbf{0.10} & \textbf{0.18} & \textbf{0.20} & 0.60 \\
		EN-ND (RD) & 0.00 & 0.00& 0.00 & 0.1\\
		\hline
	\end{tabular}
}
	\label{tbl:accuracy}
	\caption{Precision at Top-K for various language pairs. }
	\end {table}

\paragraph{Results.}
Table 4 shows the main results in terms of precision at top-k, the last column, $RD=0.10$ shows precision at the value of  $k$ which yields random chance of 10\% precision. The top  two rows show the results of bilingual dictionary induction between English and Kwanyama.  We compare the projection matrix approach, EN-KW, to  random chance, EN-KW (RD). We can see that EN-KW far outperforms chance.  This result is promising given that our  annotator only generated about $2,142$ labeled examples. In particular,   English-Italian (EN-IT) with a  larger dictionary of $5,000$ word pairs, produced by \cite{DBLP:journals/corr/DinuB14},  achieves similar numbers, however it is worth noting that the EN-IT test data set is also much larger.  For the English-Ndonga,  EN-ND, language pair, we have no labeled data. We consider three cases:  1) EN-ND (J-KW), for this case, we concatenate the Kwanyama and Ndonga corpora and use the  EN-KW training data to induce the EN-ND dictionary. 2) EN-ND (J-IT),  we  concatenate the Italian and Ndonga corpora and use the  EN-IT training data to induce EN-ND dictionary. 3) EN-ND (J-KW-R), this is our approach where we first modify the Ndonga corpus to look more like Kwanyama before  combining the two corpora, and using the  EN-KW training data.
Among these three options,  EN-ND (J-KW-R) performs  best, especially at small values of $k$, ie, k =1, 5, 10.  Additionally,  EN-ND (J-KW) outperforms EN-ND (J-IT), which is to be expected because ND, Ndonga, a Bantu language is much more similar to KW, Kwanyama than to the Indo-European language, IT, Italian.

\begin{figure}[t]
		%\centering
		\includegraphics[width=1\columnwidth]{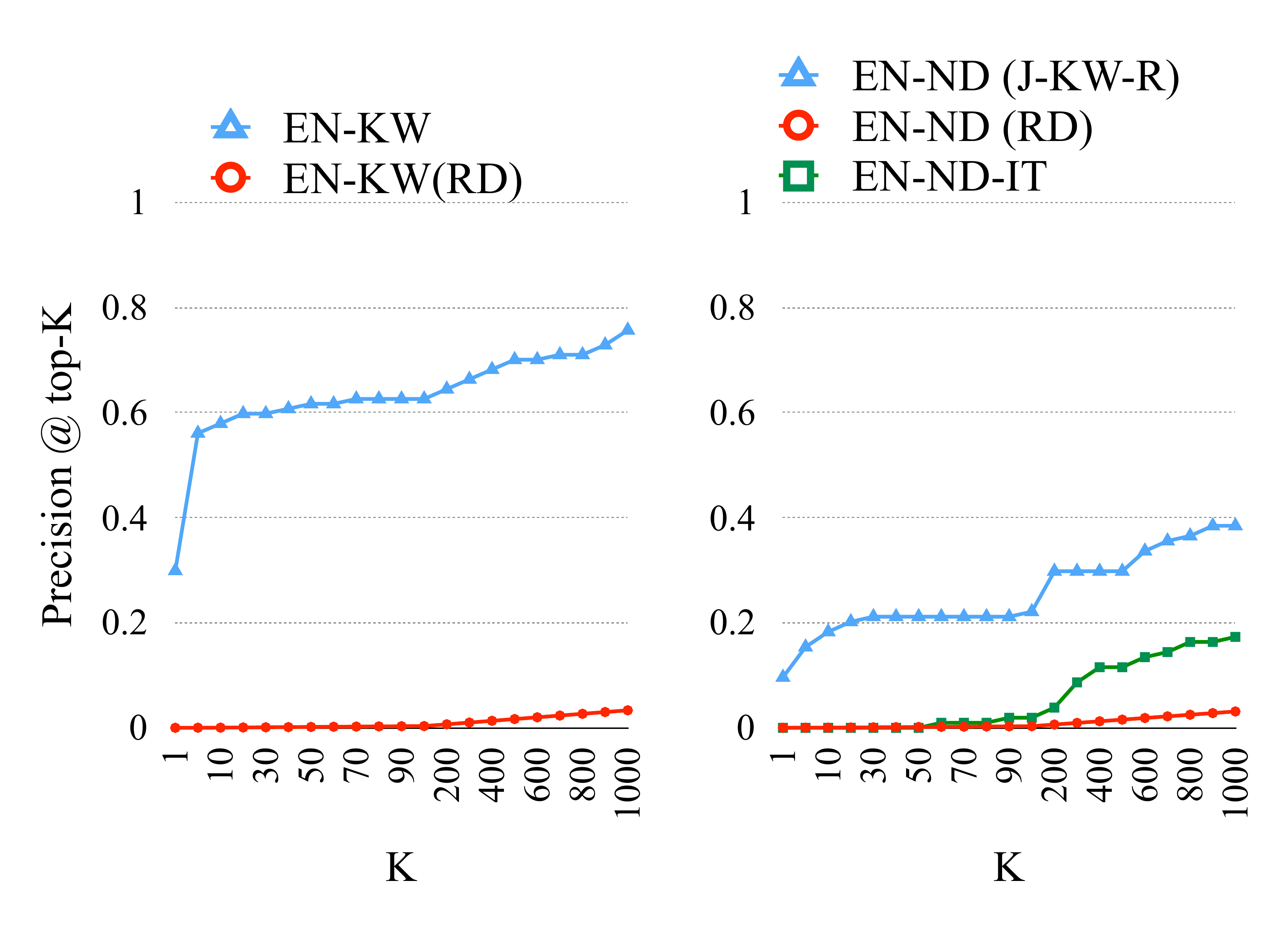}
		\vspace{-1.0cm}
		\caption{Trend for precision at top-k}
		\label{fig:trend}
	\end{figure}

Figure \ref{fig:trend} shows the top-k precision trends for  various values of $k$.
For the  EN-KW pair, left of Figure \ref{fig:trend},    there is a  bigger gap between EN-KW and  random chance EN-KW (RD).  On the other hand, for the EN-ND pair, the right of Figure \ref{fig:trend}, the  gap between our approach  EN-ND (J-KW-R) and  random choice, EN-ND (RD) is  smaller. However, it is also clear that the precision at top-k trend is much better when we make use of training data from Kwanyama  EN-ND (J-KW-R),  instead of training data from Italian  EN-ND (J-IT).
This result is encouraging for future work towards inducing accurate bilingual dictionaries  for Bantu languages  without labeled data. Future directions include collecting more training data from popular Bantu  languages such as Swahili and Zulu;  proposing alternative methods to dictionary induction; and inducing dictionaries for more Bantu languages.
	
	\section{Conclusion}
	In prior work, bilingual dictionary induction has been studied  mostly for  resource rich languages. \cite{DBLP:conf/acl/LazaridouDB15,DBLP:conf/acl/UpadhyayFDR16,DBLP:conf/eacl/FaruquiD14,DBLP:journals/corr/MikolovLS13,DBLP:journals/corr/AmmarMTLDS16,conf/acl/HaghighiLBK08}.
%	 We have introduced a method that enables dictionary induction for Bantu languages, where training data might not be available for each language. 
		We have introduced an approach where we create one vector space for Bantu languages in order to exploit labeled data available for one language but not for another. 
		Given that there are over 300 Bantu languages, and not all of them have training data, we believe  approaches that rely on  their shared grammar  will be important for bringing NLP methods to this family of languages.

\clearpage
\nocite{*}
\bibliography{dictrefs}
\bibliographystyle{acl2016}

\end{document}